\documentclass[acmlarge]{acmart}

\makeatletter
\newcommand{\confshort}{\acmConference@shortname}
\newcommand{\conffull}{\acmConference@name}
\newcommand{\confdate}{\acmConference@date}
\newcommand{\confloc}{\acmConference@venue}
\AtBeginDocument{
  \fancypagestyle{firstpagestyle}{
    \fancyhead{}%
    \fancyfoot[C]{}%
  }
  \fancyhf{}
  \fancyhead[LO]{\@headfootfont\shorttitle}%
  \fancyhead[RE]{\@headfootfont\@shortauthors}%
  \fancyhead[LE]{\@headfootfont\footnotesize \confshort, \confdate, \confloc}%
  \fancyhead[RO]{\@headfootfont\footnotesize \confshort, \confdate, \confloc}%
  \fancyfoot[C]{}%
}
\makeatother
\acmBooktitle{\conffull\@ (\confshort), \confdate, \confloc}

\usepackage{multirow}
\usepackage[table]{xcolor}
\usepackage{tikz}
\usetikzlibrary{positioning}
\AtBeginDocument{%
  }

\copyrightyear{2026}
\acmYear{2026}
\setcopyright{cc}
\setcctype{by}
\acmConference[FAccT '26]{The 2026 ACM Conference on Fairness, Accountability, and Transparency}{June 25--28, 2026}{Montreal, QC, Canada}
\acmBooktitle{The 2026 ACM Conference on Fairness, Accountability, and Transparency (FAccT '26), June 25--28, 2026, Montreal, QC, Canada}
\acmDOI{10.1145/3805689.3812341}
\acmISBN{979-8-4007-2596-8/2026/06}
\begin{document}

\title[Engaged AI Governance Through Internal Expert Collaboration]{Engaged AI Governance: Addressing the Last Mile Challenge Through Internal Expert Collaboration}
% \title{Engaged AI Governance in Technical Teams: A Collaborative Workshop Approach}

\author{Simon Jarvers}
\email{simon.jarvers@tum.de}
\orcid{0009-0003-1207-316X}
\affiliation{%
  \institution{Technical University of Munich}
  \department{Professorship for Societal Computing}
  \city{Munich}
  \country{Germany}
}

\author{Orestis Papakyriakopoulos}
\email{orestis.p@tum.de}
\orcid{0000-0003-4680-0022}
\affiliation{%
  \institution{Technical University of Munich}
  \department{Professorship for Societal Computing}
  \city{Munich}
  \country{Germany}
}

\renewcommand{\shortauthors}{Jarvers \& Papakyriakopoulos}

\begin{abstract}
Under the EU AI Act, translating AI governance requirements into software development practice remains challenging. While AI governance frameworks exist at industry and organizational levels, empirical evidence of team-level implementation is scarce. We address this ``Last Mile'' Challenge through insider action research embedded within an AI startup. We present a legal-text-to-action pipeline that translates EU AI Act requirements into actionable strategies through internal expert collaboration by extracting requirements from legal text, engaging practitioners in assessment and ideation, and prioritizing implementation through collective evaluation. Our analysis reveals three patterns in how practitioners perceive regulatory requirements: convergence (compliance aligns with development priorities), existing practice (current work already satisfies requirements), and disconnection (requirements perceived as administrative overhead). 
Based on these patterns, we discuss when governance might be treated genuinely or performatively. Practitioners prioritize requirements that serve end-users or their own development needs, but view verification-oriented requirements as box-ticking exercises. This distinction suggests a translation challenge: regulatory requirements risk superficial treatment unless practitioners understand how compliance serves system quality and user protection. Expert collaboration offers a practical mechanism for transforming governance from external imposition to shared ownership and making previously invisible governance work visible and collective.
\end{abstract}

\keywords{AI Governance, AI Regulation, EU AI Act, Participatory Design, Expert Collaboration, Implementation Research, Action Research, SMEs}

\begin{CCSXML}
<ccs2012>
   <concept>
       <concept_id>10003456.10003462.10003588.10003589</concept_id>
       <concept_desc>Social and professional topics~Governmental regulations</concept_desc>
       <concept_significance>500</concept_significance>
       </concept>
   <concept>
       <concept_id>10011007.10011074.10011134</concept_id>
       <concept_desc>Software and its engineering~Collaboration in software development</concept_desc>
       <concept_significance>500</concept_significance>
       </concept>
 </ccs2012>
\end{CCSXML}

\ccsdesc[500]{Social and professional topics~Governmental regulations}
\ccsdesc[500]{Software and its engineering~Collaboration in software development}

\maketitle

\section{Introduction}

The European landscape of AI governance is undergoing a fundamental transformation. After years of voluntary principles, the EU AI Act (Regulation 2024/1689) establishes mandatory requirements for AI systems operating in European markets. High-risk AI systems now face legally binding obligations for technical documentation, data governance, risk management, transparency, performance evaluation, human oversight, and continuous monitoring. However, regulation alone does not ensure responsible AI, as companies may evade and avoid obligations~\cite{yew2025red}, or develop a ``fine is a price'' attitude~\cite{mendez2023current}. Scholars therefore emphasize socio-technical solutions that embed ethical responsibility for all stakeholders who are involved in the design, development, and maintenance of AI systems~\cite{dignum2023responsible, papagiannidis2025responsible, radanliev2024ethics, camilleri2024artificial}.  Without such integration, the pattern observed with voluntary principles may repeat: formal compliance without meaningful change~\cite{hagendorff2020ethics, mittelstadt2019principles, jobin2019global}.

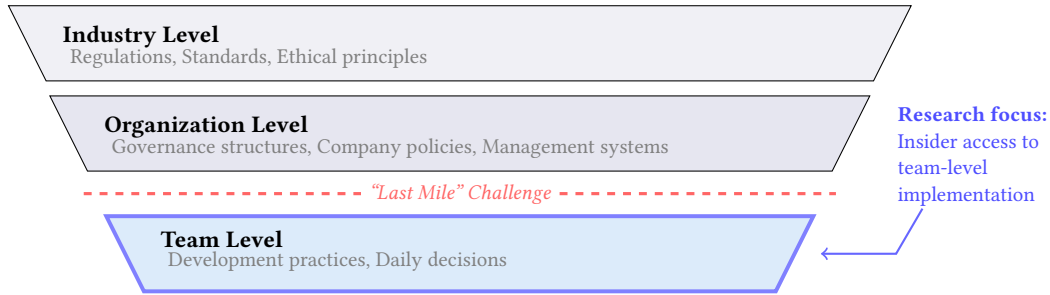
\begin{figure}[htbp]
\centering
\begin{tikzpicture}[
    font=\small
]

% Define colors
\definecolor{industrycolor}{RGB}{240,240,245}
\definecolor{orgcolor}{RGB}{230,230,240}
\definecolor{teamcolor}{RGB}{220,235,250}

% Industry Level - widest: 6 to 5.5
\coordinate (ind_tl) at (-6,5.0);
\coordinate (ind_tr) at (6,5.0);
\coordinate (ind_bl) at (-5.5,4);
\coordinate (ind_br) at (5.5,4);

\fill[industrycolor] (ind_tl) -- (ind_tr) -- (ind_br) -- (ind_bl) -- cycle;
\draw[black] (ind_tl) -- (ind_tr) -- (ind_br) -- (ind_bl) -- cycle;

\node[font=\small\bfseries, anchor=west] at (-5.4,4.6) {Industry Level};
\node[font=\footnotesize, anchor=west, text=gray] at (-5.3,4.3) {Regulations, Standards, Ethical principles};

% Organization Level - medium: 5.5 to 5
\coordinate (org_tl) at (-5.45,3.8);
\coordinate (org_tr) at (5.45,3.8);
\coordinate (org_bl) at (-4.95,2.8);
\coordinate (org_br) at (4.95,2.8);

\fill[orgcolor] (org_tl) -- (org_tr) -- (org_br) -- (org_bl) -- cycle;
\draw[black] (org_tl) -- (org_tr) -- (org_br) -- (org_bl) -- cycle;

\node[font=\small\bfseries, anchor=west] at (-4.85,3.4) {Organization Level};
\node[font=\footnotesize, anchor=west, text=gray] at (-4.75,3.1) {Governance structures, Company policies, Management systems};

% Gap visualization - positioned clearly between org and team
\draw[dashed, red!60, line width=1.2pt] (-5.0,2.5) -- (5.0,2.5);
\node[font=\footnotesize\itshape, text=red!60, fill=white, inner sep=3pt] at (0,2.5) {``Last Mile'' Challenge};

% Team Level - narrowest: 4.8 to 4.3 - highlighted
\coordinate (team_tl) at (-4.7,2.2);
\coordinate (team_tr) at (4.7,2.2);
\coordinate (team_bl) at (-4.2,1.2);
\coordinate (team_br) at (4.2,1.2);

\fill[teamcolor] (team_tl) -- (team_tr) -- (team_br) -- (team_bl) -- cycle;
\draw[blue!50, line width=1.5pt] (team_tl) -- (team_tr) -- (team_br) -- (team_bl) -- cycle;

\node[font=\small\bfseries, anchor=west] at (-4.1,1.9) {Team Level};
\node[font=\footnotesize, anchor=west, text=gray] at (-4.0,1.6) {Development practices, Daily decisions};

% Right side: Our research positioning
\draw[thick, blue!60, ->] (6.15,2.3) -- (5.8,1.7) -- (4.8,1.7);
\node[font=\footnotesize, text=blue!70, align=left, anchor=west] at (5.7,3.0) {
    % \textbf{This paper's}\\
    \textbf{Research focus:}\\
    Insider access to\\
    team-level\\
    implementation
};

\end{tikzpicture}
\caption{AI Governance Levels and the ``Last Mile'' Challenge. AI governance operates at three levels: industry (regulation and standards), organizational (policies and management systems), and team (development practices). Research concentrates on accessible upper levels through policy analysis and framework development. The transition from organizational policies to team practices presents the ``Last Mile'' Challenge, where proprietary processes and competitive concerns limit external research access~\cite{scheuerman2024walled}. Our insider action research addresses this gap through embedded access to development team processes.}
\Description[Three-level governance funnel showing the Last Mile Challenge]{A funnel diagram with three descending trapezoid layers. The top layer labeled Industry Level contains regulations, standards, and ethical principles. The middle layer labeled Organization Level contains governance structures, company policies, and management systems. The bottom layer labeled Team Level, highlighted in blue, contains development practices and daily decisions. A horizontal dashed line labeled Last Mile Challenge separates the Organization and Team levels. An arrow on the right points to the Team Level with text stating Our research: Direct access to team-level implementation.}
\label{fig:governance-levels}
\end{figure}

To understand how regulatory requirements are implemented into real-world AI systems, the process can be divided into three levels~\cite{shneiderman2020bridging, lu2024responsible}: AI governance requirements are developed at (1) the industry level through regulation and standards, adapted by companies at (2) the organizational level through policies and management systems, but must ultimately be implemented on (3) the team level by developers who build AI systems day-to-day. This last step bears the risk of superficial compliance: When governance is imposed externally rather than integrated into development workflows, practitioners may lack ownership and treat requirements accordingly. We call this disconnect between organizational policies and team-level practice \textbf{the ``Last Mile'' Challenge in AI governance}, borrowing from logistics where reaching individual destinations proves disproportionately difficult compared to establishing distribution infrastructure. 

\textbf{The core challenge} is developing methods that foster meaningful integration rather than performative compliance. Expert collaboration represents one such method. This is why we ask the research question: \textbf{How can collaborative workshops empower technical teams to integrate AI governance requirements into existing software development workflows?}

Related research has concentrated on the more accessible upper levels, analyzing regulatory frameworks and proposing ethical principles. Empirical implementation research remains scarce~\cite{birkstedt2023ai}: existing work typically interviews practitioners about challenges~\cite{ali2023walking, sloane2022german, rakova2021responsible} rather than testing solutions, and most proposed frameworks lack real-world validation~\cite{mokander2022conformity, kuehnert2025and, batool2025ai}. The team level presents particular access challenges: organizations protect proprietary processes, and competitive pressure limits transparency~\cite{scheuerman2024walled}. Studying how governance actually becomes embedded in development practice requires access that external researchers rarely obtain.

Methodologically, this paper employs insider action research to address both the access challenge and the implementation problem. Insider action research is an approach where researchers who are members of an organization study processes within that organization while actively participating in them~\cite{coghlan2007insider, coghlan2019doing}. We present a ``legal-text-to-action'' pipeline for integrating AI governance requirements into software development workflows, tested in a resource-constrained startup environment. The core of this process constitutes a collaborative workshop engaging the development team in assessing EU AI Act requirements, ideating implementation strategies, and prioritizing actions. We capture practitioners' opinions via a pre- and post-workshop survey and report the implementation of three requirements through detailed follow-up tracking of real-world actions. 

Our findings suggest that expert collaboration can surface alignments between regulatory obligations and existing development priorities. For some requirements, participants identified concrete synergies with product quality goals, while others were perceived as administrative overhead.
We discuss how practitioners' understanding of who benefits from regulatory requirements influences the quality of governance engagement, and the value of internal expert collaboration in surfacing these connections.

\begin{figure}[htbp]
\centering
\begin{tikzpicture}[
    font=\small,
    stage/.style={align=center, font=\small\bfseries},
    content/.style={align=center, font=\scriptsize, text=gray},
    arrow/.style={->, >=stealth, thick, gray!70}
]

% Draw the ellipse path (invisible, just for positioning)
\def\rx{4}  % horizontal radius
\def\ry{2.2}  % vertical radius

% Context and Purpose - entry point above the cycle
\node[stage] (context) at (-5,-0.5) {Context and Purpose};
\node[content, above=-0.05cm of context] (context-content) {\textit{Section 2}\\Organizational setting, insider positioning};

% Main cycle nodes positioned on ellipse
\node[stage] (diagnosing) at (0, -\ry*0.25) {Diagnosing};
\node[content, above=-0.05cm of diagnosing] {\textit{Section 3}\\Literature review, gap identification};

\node[stage] (planning) at (\rx*0.7, -\ry*0.7) {Planning\\action};
\node[content, right=-0.05cm of planning, anchor=west] {\textit{Section 4}\\Requirement translation,\\workshop design};

\node[stage] (taking) at (0, -\ry-0.3) {Taking action};
\node[content, below=-0.05cm of taking] {\textit{Section 5}\\Workshop execution, implementation tracking};

\node[stage] (evaluating) at (-\rx*0.7, -\ry*0.7) {Evaluating\\action};
\node[content, left=-0.05cm of evaluating, anchor=east] {\textit{Section 6}\\Strategy refinement,\\outcome discussion};

% Arrows
\draw[arrow] (context.east) to[out=0, in=180] (diagnosing.west);
\draw[arrow] (diagnosing.east) to[out=0, in=120] (planning.north);
\draw[arrow] (planning.south) to[out=-120, in=0] (taking.east);
\draw[arrow] (taking.west) to[out=180, in=-60] (evaluating.south);
\draw[arrow] (evaluating.north) to[out=60, in=180] (diagnosing.west);

\end{tikzpicture}

\caption{Action Research Cycle mapped to paper structure. Adapted from Coghlan and Brannick~\cite{coghlan2019doing}.}
\Description[Action research cycle with four phases mapped to paper sections]{A circular diagram showing four interconnected phases. Starting at the top right and moving clockwise: Diagnosing (Section 3: Literature review, gap identification) connects to Planning action (Section 4: Requirement translation, workshop design) connects to Taking action (Section 5: Workshop execution, implementation tracking) connects to Evaluating action (Section 6: Strategy refinement, outcome discussion) which connects back to Diagnosing. At the top left hand corner, Context and Purpose (Section 2: Organizational and insider positioning) feeds into the cycle.}
\label{fig:action-research-cycle}
\end{figure}
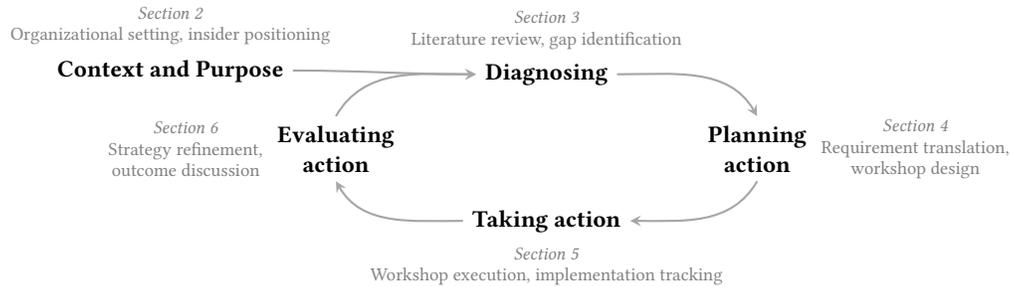

The structure of the paper follows the action research cycle of diagnosing, planning action, taking action, and evaluating action, as illustrated in Figure~\ref{fig:action-research-cycle}.

\section{Context and Purpose}
\label{sec:context}

\textbf{Research Context.} The research was conducted at an AI startup developing training and execution support solutions for industrial environments. The company employs approximately ten people in development and related roles, holds ISO 27001 and ISO 42001 certifications, and while current products do not fall into EU AI Act high-risk categories, leadership decided to prepare proactively for potential future requirements.

The first author holds dual roles as (1) AI Governance Officer responsible for developing and implementing governance frameworks, and (2) researcher investigating AI governance implementation. This arrangement provides direct access to governance processes as they unfold.

\textbf{Organizational Governance Context.} The company's ISO 42001\footnote{ISO/IEC 42001:2023 is the International Standard for AI management systems. It is not officially recognized as a harmonized European Standard and therefore not sufficient for complete high-risk EU AI Act compliance. For more information, see \url{https://www.iso.org/home/insights-news/resources/iso-42001-explained-what-it-is.html}} certification establishes organizational-level governance structures: policies, procedures, and commitments to develop and use AI responsibly. However, a management system standard and product safety law operate at different levels. ISO 42001 certifies that an \textit{organization} has processes for governing AI, analogous to how ISO 9001 certifies quality management without specifying what any particular product must do. The EU AI Act, by contrast, mandates requirements per \textit{AI system}: what it must log, how it must be documented, what transparency it must provide. Holding organizational certification does not automatically translate into team-level awareness of these product-level requirements. This disconnect is the ``Last Mile'' gap our research investigates.

\textbf{The Dual Purpose of Action Research.} Action research combines problem solving with knowledge generation. Coghlan and Brannick define it as ``an approach to research which is based on a collaborative problem-solving relationship between researcher and client which aims at both solving a problem and generating new knowledge''~\cite{coghlan2019doing}. Both dimensions are present in our study.

The problem-solving dimension addresses a practical challenge: making AI governance actionable within a resource-constrained Small and Medium-sized Enterprise (SME) and integrating it into existing software development workflows rather than maintaining it as a separate compliance function. We do not claim our approach achieves complete AI Act compliance nor do we want to present a framework for an AI Act conformity assessment. Rather, we seek a workable path toward compliance given typical SME constraints.

The knowledge-generation dimension produces insights extending beyond the immediate organizational context. Through systematic evaluation and reflection, we document how collaborative engagement functions as a mechanism for governance integration. As Coghlan and Brannick note, ``the value in action research is not whether the change process was successful or not, but rather that the exploration of the data provides useful and interesting theory''~\cite{coghlan2019doing}.

\textbf{Managing Insider Positioning.} The insider position creates tensions between organizational responsibilities and research objectives. Coghlan~\cite{coghlan2007insider} identifies three challenges: preunderstanding (assumptions from organizational experience that may create blind spots), role duality (managing organizational membership alongside research), and organizational politics (navigating confidentiality and career considerations).

We address these through several mechanisms. An external researcher co-facilitated the workshop, providing independent observation. Academic supervision ensures methodological decisions are reviewed independently of operational pressures. The research objective does not conflict with company interests: the goal is investigating whether expert collaboration enhance governance practices, not demonstrating that compliance has been achieved. Findings revealing challenges or limitations are as valuable as findings showing success.

% Despite these tensions, insider positioning enables direct participation in governance implementation as it occurs, and longitudinal tracking of which strategies are implemented, what barriers emerge, and whether initial engagement translates to sustained practice change.

\section{Diagnosing}
\label{sec:background}

Following the action research cycle, this section diagnoses the landscape of AI governance implementation to identify gaps that motivate our intervention.

\subsection{The Challenge of Translating Regulations into Actions}
\label{subsec:translating-challenge}

Translating regulatory requirements into development practice remains a persistent challenge. Academic work has produced valuable frameworks addressing this translation at different levels. Shneiderman~\cite{shneiderman2020bridging} proposed a three-layer governance structure with recommendations across team, organization, and industry levels. Lu et al.~\cite{lu2024responsible} developed patterns operationalizing ethics principles into governance, process, and design categories. Addressing the EU AI Act specifically, co-design approaches have produced impact assessment templates~\cite{bogucka2024co} and justice-oriented toolkits through expert collaboration~\cite{hollanek2025toolkit}.
These efforts engage compliance experts or cross-sectoral stakeholders in creating tools \textit{for} development teams. Our work complements this by empirically studying what happens \textit{when} an existing development team engages with regulatory requirements directly, capturing practitioner perceptions and prioritization dynamics during team-level operationalization.

Beyond academic frameworks, official and practitioner resources provide additional guidance. The EU AI Act Service Desk offers compliance checkers and requirement explorers, though these operate at the policy interpretation level rather than offering team-level implementation guidance. At the time of this study the official harmonized European Standards that would provide concrete technical specifications remain under development. ISO 42001, a globally recognized AI Management System standard, offers organizational-level structure but is not approved by the EU AI Office as a harmonized standard for AI Act compliance. Practitioner resources such as the AppliedAI white paper on AI Act governance\footnote{\url{https://www.appliedai.de/en/insights/ai-act-governance-best-practices-for-implementing-the-eu-ai-act/}} provide practical orientation, though evaluating and adapting such resources creates its own burden for resource-constrained organizations.

A gap persists between these resources and implementation. Much analysis remains at the level of organizational recommendations without attention to how requirements integrate into development workflows. When tools are proposed, empirical validation of their effectiveness in actual development contexts is rare~\cite{birkstedt2023ai, kuehnert2025and}. The challenge is not lack of frameworks but lack of evidence about how they function when teams actually use them.

More recently, the ``bridging the gap'' metaphor itself has been challenged. Ruster and Davis~\cite{ruster2025gaps} argue through fieldwork with three AI startups that principles and practices are not separated by a chasm but are ``integrated, nonlinear, and dynamically evolving.'' We support this distinction: governance implementation is not a one-time crossing but an ongoing process of negotiation. Our intervention tests one mechanism for supporting this ongoing negotiation.

\subsection{Internal Expert Collaboration versus External Participation}

Participation is a significant topic in FAccT research. The ``participatory turn'' in AI design has produced frameworks mapping modes of participation to goals, scope, and forms of engagement~\cite{delgado2023participatory, birhane2022power, kallina2024stakeholder} and has recently also been examined in the context of the EU AI Act~\cite{ullstein2025participatory}. However, terminological ambiguity obscures important distinctions. Kallina et al.~\cite{kallina2025stakeholder} argue that clearer terminology is ``crucial to avoid confusion and `ethics washing','' proposing distinctions among others between participatory development (involving affected communities), and expert involvement (consulting domain experts). Most FAccT research on participation focuses on external stakeholder participation: engaging end-users, affected communities, and civil society in shaping AI systems. The power dynamic concerns those potentially harmed by AI gaining voice in decisions affecting them. This work addresses democratic legitimacy and the redistribution of decision-making power. Studies find that such external participation faces structural tensions with commercial priorities, with stakeholder involvement often driven by business interests rather than justice concerns~\cite{kallina2025stakeholder}.

Our focus differs. We investigate internal \textit{expert collaboration} in implementing AI governance requirements, where participants are development team members. Following Kallina et al.'s terminology, this constitutes expert involvement, though we extend this concept to collaboration as participants actively co-design governance strategies. The goal is implementation effectiveness and distributed ownership rather than democratic legitimacy.

This approach draws from Scandinavian participatory design traditions, which emerged in workplace contexts to involve workers in designing technologies affecting their labor~\cite{asaro2000transforming}. In our case, the affected stakeholders are software engineers who are expected to adhere to the AI governance frameworks. Three benefits emerge from the literature for this approach: developers possess domain knowledge about workflows that governance officers may lack~\cite{nahar2022collaboration}; collaborative strategies face less resistance than imposed requirements~\cite{ali2023walking}; and the collaborative process surfaces assumptions and creates shared understanding~\cite{deng2023investigating}.

We do not suggest that internal \textit{expert collaboration} is sufficient for responsible AI development. External stakeholder engagement addresses concerns internal teams cannot: ensuring affected communities have voice, identifying harms invisible to developers, and providing democratic accountability. Internal \textit{expert collaboration} addresses a different problem: ensuring governance becomes embedded in practice rather than remaining a superficial documentation and box-ticking exercise.

\subsection{Diagnosis}

The preceding analysis motivates our intervention along both dimensions of action research.

\textbf{Problem-solving Dimension}. Despite valuable frameworks for translating legal requirements into organizational practice, a persistent challenge remains: making governance actionable within development teams given real-world constraints. In the remaining paper we present how we developed and evaluated one potential solution to this challenge suitable to our context: a focused workshop format that creates explicit space for bridging work---the iterative labor required to translate between different stakeholders' perspectives---seeks alignment between regulatory requirements and existing development priorities, and fits within the time and resource constraints of a small organization.

\textbf{Knowledge-generation Dimension}. Three gaps in current understanding motivate our research. First, while frameworks exist, empirical evidence of how governance integration actually unfolds when development teams engage with requirements remains scarce. Second, team-level implementation is under-researched because access is difficult; our insider positioning addresses this directly. Third, participatory approaches in AI governance have focused primarily on external stakeholder engagement; how internal \textit{expert collaboration} functions as a mechanism for governance integration is less understood.

\section{Planning Action}
\label{sec:methods}

\begin{figure}[t]
\centering
\includegraphics[width=\textwidth]{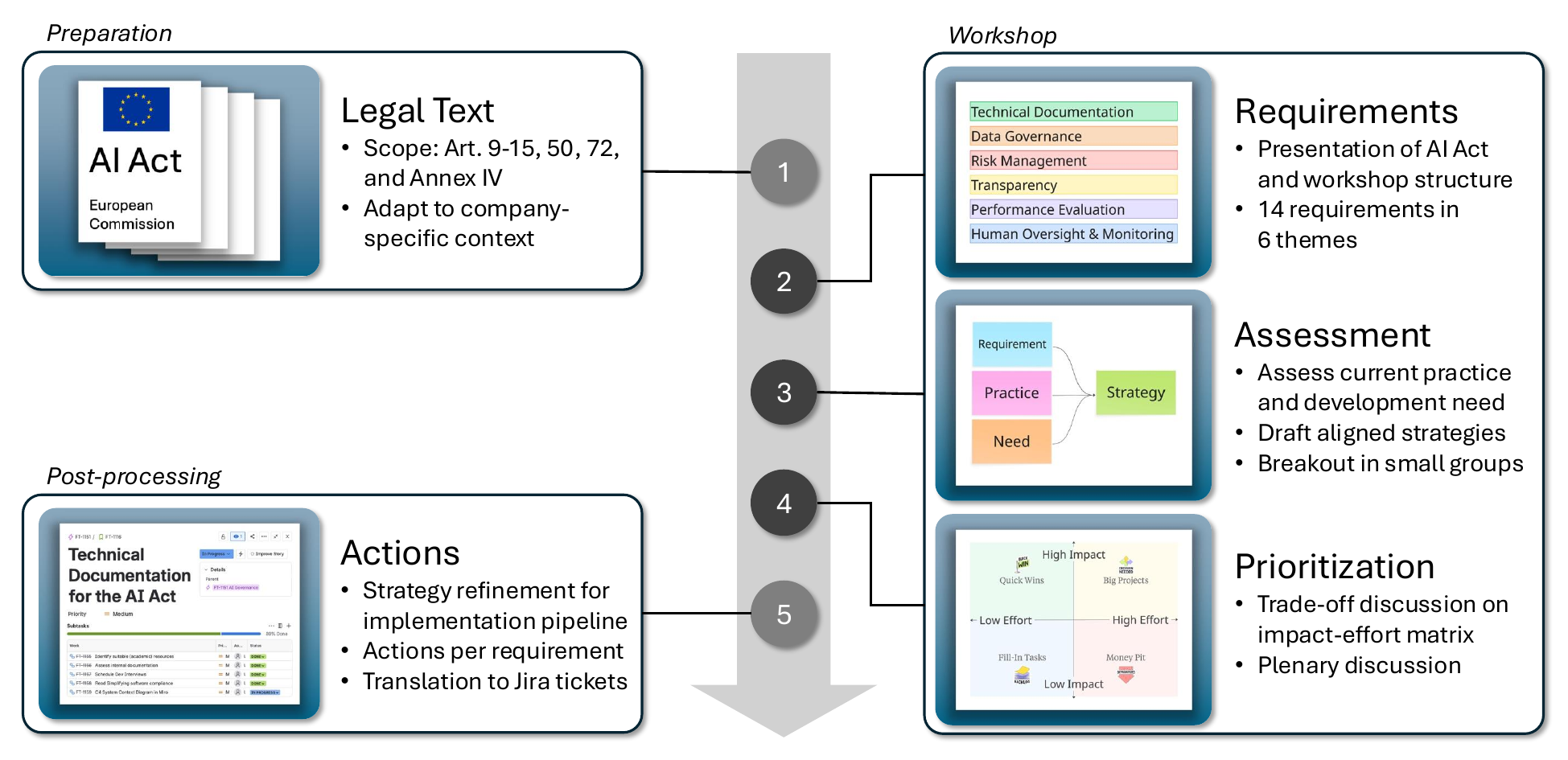}
\caption{From legal text to action: A translation pipeline for operationalizing EU AI Act requirements. The collaborative workshop forms the core of this pipeline, where practitioners learn about AI Act requirements, brainstorm implementation strategies grounded in current practices and development needs, and prioritize generated strategies based on impact and effort.}
\Description[Workshop pipeline: three phases with five sequential steps]{A workflow diagram organized in three sections. Left upper section labeled Preparation contains step 1 showing an EU AI Act document icon. Right section labeled Workshop contains three steps: step 2 shows a box with colored bars representing six themes (Technical Documentation, Data Governance, Risk Management, Transparency, Performance Evaluation, Human Oversight and Monitoring); step 3 displays a network diagram with four connected boxes showing requirement, practice, need feed into the strategy post-it; step 4 presents a box with impact-effort matrix quadrants labeled Quick Win, Big Project, Fill-In Tasks, and Money Pit. Left lower section labeled Post-processing contains step 5 showing a screenshot of a project management board. In the middle are circles with the five steps sequentially from 1 through 5, from top to bottom.}
\label{fig:pipeline}
\end{figure}

Translating legislation into team-level practices presents a core challenge for AI governance implementation. This section describes our approach to that transformation. Figure~\ref{fig:pipeline} illustrates the pipeline from legal text to actionable governance practices through five steps: preparation, presentation, assessment, prioritization, and post-processing. The collaborative workshop sits at the center of this pipeline, engaging the development team in translating requirements into strategies aligned with their existing work.

\subsection{From Legal Text to Requirements}

Given the limitations of existing resources outlined in Section~\ref{subsec:translating-challenge}, we worked directly with the legal text. After conducting an AI Act risk classification, we assessed Articles 9--15, 50, 72, and Annex IV as most relevant from the developer perspective. Building on prior work by the first author, which included consultation with legal scholars to address interpretive ambiguities, we extracted and consolidated the relevant provisions into specific requirements. The scope was deliberately bounded: the goal was not to produce an exhaustive conformity assessment, but to identify requirements that could meaningfully inform development team practices. Requirements primarily concerning organizational leadership (governance structures, resource allocation) or external verification (conformity assessment, market surveillance) were set aside for separate treatment.

This extraction produced 14 high-level requirements organized into six thematic pillars: Technical Documentation, Data Governance, Risk Management, Transparency, Performance Evaluation, and Human Oversight \& Monitoring. The framework and research design was reviewed and validated through weekly meetings with the second author, a senior researcher independent of the company.

Two preparatory meetings with the CTO adapted these requirements to the company context. The first meeting established strategic alignment: confirming company priorities for proactive governance preparation, securing approval for the workshop and its research component, and agreeing on the involvement of an external co-facilitator. The second meeting reviewed the adapted requirements to ensure relevance to actual development work, finalized participant selection, and settled operational details.

The 14 requirements from this preparation provided the foundation for the workshop.

\subsection{Workshop Design}

The workshop structure emerged from consultation with four researchers with backgrounds in political and computer science that are experienced in participatory research and conducting workshops. These consultations informed a three-phase structure following the diamond model of divergent-then-convergent thinking: first expanding the solution space through open ideation (step 2), then narrowing toward concrete priorities through structured evaluation (step 3).

\begin{enumerate}
    \item \textbf{Presentation.} The first author presented the AI Act context, explained the governance objectives motivating the workshop, and introduced the 14 identified requirements. This phase established shared vocabulary and ensured all participants understood the context and purpose of the workshop.
    \item \textbf{Assessment and Ideation.} Participants were purposefully split into three groups to balance their domain knowledge and seniority level. Each group got assigned to two pillars. They documented their current practices for each requirement, identified product and development needs, and generated implementation strategies. The format encouraged practitioners to draw on their implicit knowledge of existing workflows, something a governance officer working in isolation would lack.
    \item \textbf{Prioritization.} Groups reconvened for plenary discussion. Each group presented their strategies, followed by collective mapping onto an impact-effort matrix. This phase surfaced trade-offs, built consensus on priorities, and ensured the resulting action plan reflected team judgment rather than top-down imposition.
\end{enumerate}

The workshop was supported by an external co-facilitator. Within a preliminary meeting with the external facilitator, the workshop design was tested, and the distribution of roles were established: the first author led content presentation and requirement explanation while the co-facilitator supported group work and documented observations from an independent perspective.

\textbf{Participants and Consent.} The workshop was conducted in November 2025, lasting 90 minutes. This was the maximum resource investment from an economic perspective of the leadership. Eight participants attended: three AI engineers, four full-stack developers, and a product owner. The format was hybrid, with two participants joining remotely.
A participant information sheet was distributed three days before the workshop, explaining both the operational purpose (developing governance strategies) and the research component (studying expert collaboration). Workshop participation was a company activity during paid working hours. Contribution of data to the research was voluntary, with informed consent obtained and documented within the pre-workshop survey.
% Moved to Ethics statement
% Given the small sample size and the first author's insider position, complete anonymity for survey responses was not possible. We therefore adopted confidentiality-based protocols: individual survey responses would not be shared with the employer, only aggregated findings would be reported, and publications would use pseudonyms where appropriate. The workshop itself, of course, involved face-to-face discussion among colleagues and could not be anonymous.

\textbf{Data Collection.} Data was collected during different stages: a \textit{pre- and post-workshop survey} was distributed among participants to capture their attitudes, knowledge, and opinions of the workshop. Given the small sample size (n=8), we focus on descriptive analysis to inform our interpretation. During the workshop, a \textit{collaborative, digital whiteboard} (i.e., Miro board) was used to document status quo assessments, identified needs, generated strategies, and prioritization outcomes. The external co-facilitator documented \textit{observations} independently, focusing on group dynamics, moments of insight or resistance, and discussion patterns that the first author might have missed while facilitating content. An overview of survey questions and the observation notes template can be found in Appendix~\ref{app:data}.

% Data collection served the dual purposes of action research: generating evidence for the knowledge-generation dimension while supporting the problem-solving dimension through systematic documentation.

% \textbf{Surveys.} Pre-workshop surveys (distributed three days before) captured baseline attitudes toward AI governance, self-assessed knowledge of the EU AI Act and ISO 42001, and open-ended associations with ``AI governance.'' Post-workshop surveys (administered immediately after the workshop) repeated the attitude measures, asked for concrete examples of perceived compliance-quality alignment, assessed workshop value, and captured preferences for governance responsibility distribution. Consistent Likert scales enabled pre-post comparison. Given the small sample size (n=8), we focus on descriptive analysis to inform our interpretation.

% \textbf{Workshop artifacts.} The Miro board captured status quo assessments, identified needs, generated strategies, and prioritization outcomes. These artifacts document both the substance of team discussion and the reasoning behind prioritization decisions.

% \textbf{Observation notes.} The external co-facilitator documented observations independently, focusing on group dynamics, moments of insight or resistance, and discussion patterns that the first author might have missed while facilitating content.

\subsection{Post-Processing}

Following the workshop, the generated strategies required translation into actionable form. The authors assessed workshop artifacts, refined the strategies based on prioritization outcomes and feasibility considerations, and developed concrete actions corresponding to each of the 14 requirements.

These refined strategies and actions were discussed with the company's CTO to establish a 12-month implementation roadmap. Concrete actions were translated into tickets within the company's project management system (Jira), integrating governance work into the normal development workflow rather than maintaining it as a separate track.

Three of these concrete actions are reported in detail in Appendix~\ref{app:tracking}: implementing a self-hosted AI monitoring system (i.e., Langfuse) to address logging and performance evaluation requirements, auditing AI interaction disclosure to address transparency requirements, and establishing technical documentation practices to address documentation requirements. These examples illustrate how workshop-generated strategies translated into actual development work.

\section{Taking Action}
\label{sec:results}

The workshop produced implementation strategies for all 14 EU AI Act requirements across the six governance pillars. Table~\ref{tab:workshop-outcomes} presents the complete mapping from requirements to concrete actions. This section analyzes these outcomes to identify patterns in how practitioners engage with different types of requirements, drawing on follow-up tracking of selected implementations to ground our interpretation.

\begin{table*}[htbp]
    \centering
    \renewcommand{\arraystretch}{1.2}
    \setlength{\tabcolsep}{2.5pt}
    \small
    \begin{tabular}{>{\raggedright}p{1.6cm} 
                    >{\raggedright}p{0.4cm} 
                    >{\raggedright}p{1.8cm}
                    >{\raggedright}p{1.0cm}
                    >{\raggedright}p{9.5cm}
                    >{\raggedright\arraybackslash}p{0.6cm}}
        \hline
        \textbf{Pillar} & \textbf{ID} & \textbf{Requirem.} & \textbf{Article} & \textbf{Actions (First Actions in Bold)} & \textbf{Pat.} \\
        \toprule
        
        \cellcolor[HTML]{B9F1D1} Technical Documentation & 
        1.1 & 
        Technical Documentation & 
        11; IV & 
        \textbf{Develop C4 architecture diagrams.} Use simplified technical documentation for SMEs (to be released by EC); assign ownership; consolidate existing documentation; clarify maintenance strategy & 3 \\
        \hline
        
        \cellcolor[HTML]{FADBBD} Data Governance & 
        2.1 & 
        Data Governance & 
        10 & 
        \textbf{Create data flow diagram.} Clarify scope for foundation-model users (testing data, input quality, output lineage); structure test data catalog & 3 \\
        \hline
        
        \cellcolor[HTML]{FFD2D0} Risk Management & 
        3.1 & 
        Risk Management & 
        9 & 
        \textbf{Update existing risk management system.} Communicate ISO 42001 risk management to team; connect monitoring data to risk identification & 3 \\
        \hline
        
        \cellcolor[HTML]{FFF6C3} & 
        4.1 & 
        Changes Disclosure & 
        13(3c); IV(2f)(6) & 
        \textbf{Evaluate channel options (trust center, email, in-app).} Formalize release notes template; integrate performance metrics from monitoring & 1 \\
        \cline{2-6}
        
        \cellcolor[HTML]{FFF6C3} & 
        4.2 & 
        Interpretable System Design & 
        13(1) & 
        \textbf{Update API documentation.} Two-track approach: (A) Improve API documentation for technical deployers; (B) Add end-user interpretability features (quality indicators, content origin marking) & 1 \& 2 \\
        \cline{2-6}
        
        \cellcolor[HTML]{FFF6C3} Trans- \\ parency & 
        4.3 & 
        AI Interaction Disclosure & 
        50(1) & 
        \textbf{Audit current AI disclosure touchpoints.} Standardize AI disclosure across platform; create UX pattern for consistency; ensure accessibility & 2 \\
        \cline{2-6}
        
        \cellcolor[HTML]{FFF6C3} & 
        4.4 & 
        Synthetic Content Marking & 
        50(2) & 
        \textbf{Design content metadata schema (origin, timestamp, sources, edit history).} Implement content provenance system: metadata tagging at creation; visual indicators for AI content; version control for edit tracking & 1 \\
        \hline
        
        \cellcolor[HTML]{E5E2FF} & 
        5.1 & 
        Logging System & 
        12 & 
        \textbf{Set up Langfuse instance.} Deploy Langfuse as central governance infrastructure serving logging, metrics, monitoring, and debugging needs & 1 \\
        \cline{2-6}
        
        \cellcolor[HTML]{E5E2FF} \vspace{1.5mm} Performance & 
        5.2 & 
        Metrics \& Validation & 
        15(1)(3); IV & 
        \textbf{Document existing evaluation framework.} Expand test suite with customer data; leverage Langfuse; establish human evaluation protocol & 1 \& 2 \\
        \cline{2-6}
        
        \cellcolor[HTML]{E5E2FF} Evaluation & 
        5.3 & 
        Robustness \& Resilience & 
        15(4) & 
        \textbf{Document existing robustness measures.} Update error handling based on monitored data; conduct stress testing; develop custom deployment & 1 \& 2 \\
        \cline{2-6}
        
        \cellcolor[HTML]{E5E2FF} & 
        5.4 & 
        Cybersecurity Measures & 
        15(5); IV(2h) & 
        \textbf{Document security measures aligned with ISO 27001.} Layered security: input scanning, AI threat model, regular penetration testing & 2 \\
        \hline
        
        \cellcolor[HTML]{D0E3FF} \vspace{4.8mm} Human & 
        6.1 & 
        Human Oversight Design & 
        14 & 
        \textbf{Create human oversight documentation with screenshots and workflow diagrams.} Document current oversight design (reviewing, access controls, no autonomous actions); design automation bias warnings & 2 \\
        \cline{2-6}
        
        \cellcolor[HTML]{D0E3FF} Oversight \& Monitoring & 
        6.2 & 
        Instructions for Use & 
        13(2)(3); 14(4) & 
        \textbf{Review and update existing Instructions for Use document.} Populate with current information; introduce tutorial feature for in-app guidance & 1 \& 2 \\
        \cline{2-6}
        
        \cellcolor[HTML]{D0E3FF} & 
        6.3 & 
        Post-Market Monitoring & 
        72; 9(2c) & 
        \textbf{Implement post-market monitoring plan based on EC template (not yet available).} Leverage Langfuse; formalize customer feedback & 1 \\
        \bottomrule
    \end{tabular}
    \caption{Refined workshop outcomes: 14 EU AI Act requirements mapped to actions across 6 governance pillars. Included Articles of the AI Act are: 9--15, 50, 72, and Annex IV. The Workshop was conducted with 8 development team members (3 AI engineers, 4 full-stack developers, 1 product person).}
    \label{tab:workshop-outcomes}
\end{table*}

\subsection{Three Patterns of Requirement Engagement}

Analyzing the workshop outcomes and their subsequent implementation reveals three distinct patterns in how development teams relate to regulatory requirements. These patterns can help to understand how practitioners' interpretation of requirements shapes governance engagement.

\subsubsection{Pattern 1: Convergence Between Compliance and Quality Goals}

Some requirements address concerns that development teams already prioritize. Here, regulatory obligations and development needs converge on shared solutions, creating genuine alignment rather than imposed burden.
\textit{Logging System} (Req. 5.1) exemplify this pattern. Article 12 mandates automatic recording of events relevant for risk identification and post-market monitoring. For our development team, this requirement converged with an existing priority: implementing observability tools for debugging AI behavior and diagnosing customer-reported errors. The team implemented Langfuse, an open-source observability platform, to address both needs through a single solution. One AI engineer captured the alignment: ``Previously when customers reported errors, we relied on their descriptions which were often incomplete. Now we can trace exactly what happened.'' A workshop participant made the connection explicit: ``Logging requirement directly aligns with our drive to improve quality through an understanding of system steps.'' The compliance requirement did not impose new work but validated and structured work the team already wanted to do.
Other requirements showing this convergence pattern include \textit{Changes Disclosure} (Req. 4.1), which aligned with needs to improve customer communication, and \textit{Instructions for Use} (Req. 6.2), which connected to ongoing efforts supporting deployers navigating a growing feature set.

\subsubsection{Pattern 2: Requirements Satisfied by Existing Practice}

Where Pattern 1 involves new implementation work motivated by shared goals, Pattern 2 captures cases where existing practice already satisfies the requirement with minimal additional effort.
These requirements formalize what teams already do, often for reasons predating any regulatory consideration. 
\textit{AI Interaction Disclosure} (Req. 4.3) exemplifies this pattern. Article 50(1) requires that natural persons interacting with an AI system are informed unless obvious from context. The workshop identified this as a ``quick win'' requiring verification rather than new development. The subsequent audit confirmed that existing design choices---feature naming, iconography, and introductory messaging---already satisfied the requirement. Compliance work consisted solely of documenting these existing practices.
This pattern appeared across several requirements. \textit{Human Oversight Design} (Req. 6.1) were built into the product as deliberate UX decisions: creating what one developer called a ``feeling of co-creation'' rather than outsourcing tasks to AI entirely. \textit{Cybersecurity Measures} (Req. 5.4) aligned with existing ISO 27001 certification maintained for business reasons.
The workshop revealed that teams were already ``doing governance'' without recognizing it as compliance, a finding consistent with prior research~\cite{koh2024voices}. The collaborative format surfaced these existing practices and connected them to regulatory language.

\subsubsection{Pattern 3: Requirements Perceived as Disconnected from Practice}

Unlike the previous two patterns, where practitioners either see shared goals (Pattern 1) or recognize existing practice (Pattern 2), some requirements add work where practitioners see no clear connection to system quality or user experience. Here, compliance risks becoming symbolic: documentation satisfying formal criteria without genuine engagement.
The perception of disconnection was consistent across these requirements, though the underlying reasons differed: some requirements were perceived as burdensome paperwork and rigid processes competing for scarce resources in fast-paced development environments, while others faced legal ambiguity that left practitioners uncertain about what compliance entails.

\begin{figure}[t]
\centering
\includegraphics[width=0.6\textwidth]{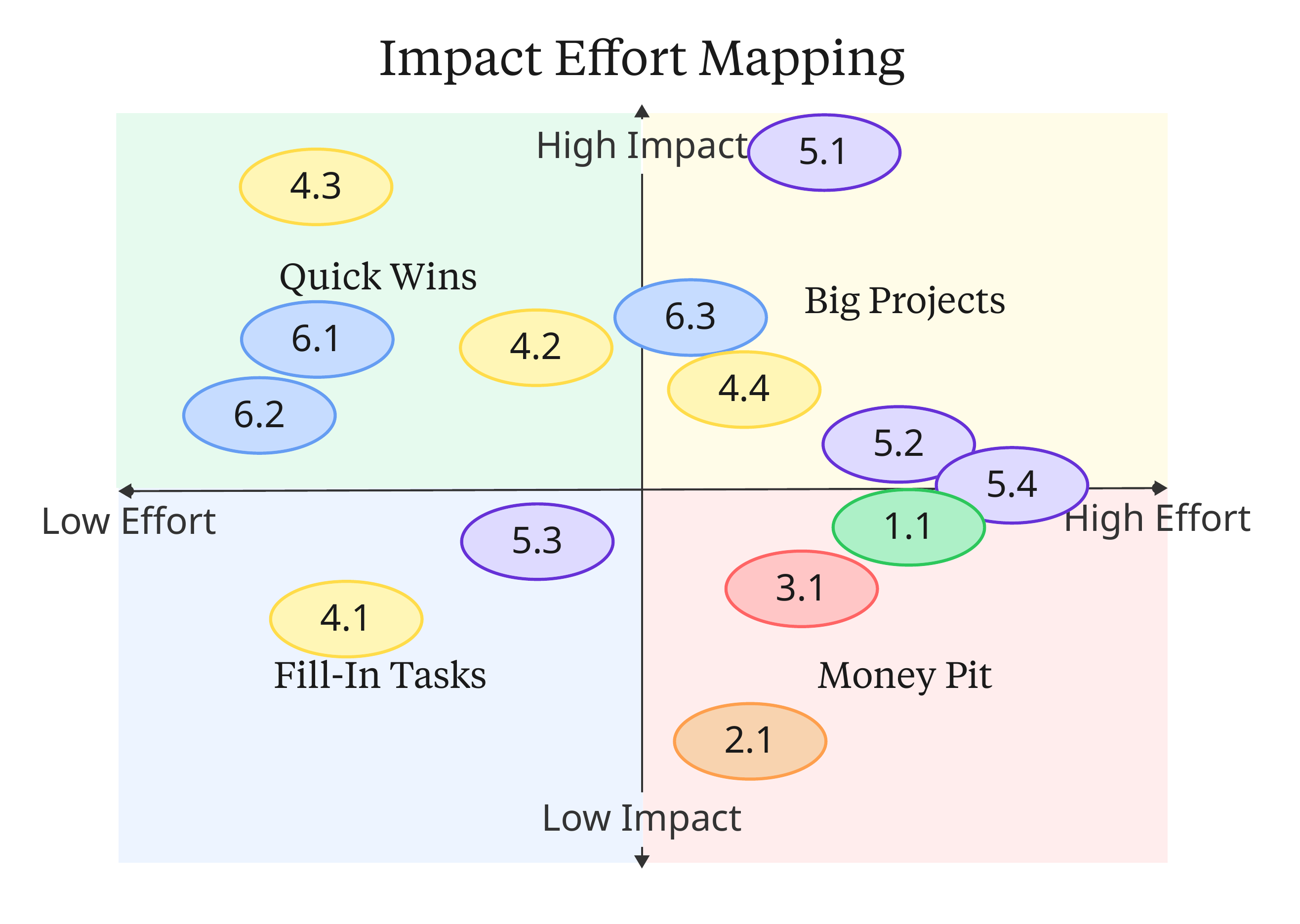}
\caption{Impact-effort mapping of the 14 workshop-generated strategies. Participants discussed each strategy until reaching consensus on placement within the continuous matrix; this distribution emerged from practitioners' own assessments rather than facilitator framing. Numbers and colors correspond to Table~\ref{tab:workshop-outcomes}. Notable clustering emerged: \textcolor[HTML]{fbe16a}{\textbf{Transparency (4.X)}} and \textcolor[HTML]{a4c0e7}{\textbf{Human Oversight \& Monitoring (6.X)}} requirements fell within or near the Quick Wins quadrant (high impact, low effort), \textcolor[HTML]{a296e8}{\textbf{Performance Evaluation (5.X)}} requirements clustered in the Big Projects quadrant (high impact, high effort), while \textcolor[HTML]{45be77}{\textbf{Technical Documentation (1.X)}}, \textcolor[HTML]{feaf62}{\textbf{Data Governance (2.X)}}, and \textcolor[HTML]{ffaba9}{\textbf{Risk Management (3.X)}} landed in the Money Pit quadrant (low impact, high effort).}
\Description[Impact-effort matrix showing distribution of 14 workshop strategies]{A two-by-two matrix with Low Effort to High Effort on the horizontal axis and Low Impact to High Impact on the vertical axis, creating four quadrants. Top-left quadrant ``Quick-Wins'' in light green contains four strategy bubbles labeled 4.3, 6.1, 6.2, and 4.2. Top-right quadrant ``Big Projects'' in yellow contains five strategy bubbles labeled 6.3, 4.4, 5.1, 5.2, and 5.4. Bottom-left quadrant ``Fill-in Tasks'' in light blue contains two strategy bubbles labeled 4.1 and 5.3. Bottom-right quadrant ``Money Pit'' in light red contains three strategy bubbles labeled 3.1, 1.1 and 2.1. Strategy bubbles vary in color (yellow, blue, purple, green, red, and orange) indicating different thematic categories from the six-pillar framework.}
\label{fig:impact-effort}
\end{figure}

\textit{Technical Documentation} (Req. 1.1) exemplifies
the burden dynamic.
Article 11 requires comprehensive technical documentation as specified in Annex IV. When asked about benefits beyond compliance, developers assessed the value as ``negligible.'' While documentation could theoretically support onboarding or customer communication, these potential benefits were not perceived as sufficient to justify the effort independently of compliance requirements. The ongoing maintenance burden was identified as a significant concern.
\textit{Risk Management} (Req. 3.1) followed a similar logic. Developers engage in intuitive risk identification and mitigation as part of their daily work when debugging, testing, and reviewing code. However, they perceived the formal and auditable risk management process mandated by Article 9 as rigid and disconnected from these existing practices. The requirement landed in the ``Money Pit'' quadrant (Figure~\ref{fig:impact-effort}), because the formalized process was perceived as serving auditors rather than informing their own work.

\textit{Data Governance} (Req. 2.1) and the delayed SME documentation template for Technical Documentation illustrate the ambiguity dynamic. Article 11(2) mandates a simplified documentation template for SMEs, but eighteen months after enactment this template remains unavailable, leaving SMEs uncertain whether to invest in comprehensive documentation now or wait for anticipated simplification. Similarly, as foundation model users who do not train their own models, the team struggled to identify applicable actions
for data governance.
The EU AI Act as horizontal regulation addresses diverse AI systems through uniform requirements, but this generality creates uncertainty for specific use cases.
Here, the perceived disconnection stems from legal ambiguity.

\subsection{The Risk of Performative Compliance}

These three patterns document how practitioners' interpretation of regulatory requirements shapes whether they might treat governance genuinely or performatively. The distinction hinges on a question that emerged organically during the workshop: \textit{Who benefits from this requirement?}

\textbf{Cui bono?} The impact-effort prioritization (Figure~\ref{fig:impact-effort}) made this dynamic visible. Requirements serving end-users (AI interaction disclosure, human oversight, instructions for use) or developers (logging, performance monitoring) clustered as high-impact. Requirements serving primarily auditors and regulators (technical documentation, formal risk management) clustered as low-impact despite substantial effort. This distribution emerged from practitioners' own assessments, not from facilitator framing.

The pattern suggests a simple logic: software developers care about making products work well. When regulatory requirements connect to this professional commitment, practitioners engage meaningfully. When requirements are perceived as serving external verification rather than system improvement, they receive correspondingly superficial treatment. Compliance experienced as box-ticking will be approached as box-ticking.

\textbf{Implications for Regulatory Design.} This observation does not argue that verification-oriented requirements are unnecessary; external oversight of self-interested organizations requires evidence. However, the same legal obligation can be understood as ``document your system so you understand it and can improve it'' or as ``document your system so external parties can verify what you did.'' Both framings may be accurate, but practitioners in our workshop engaged genuinely with the former framing and performatively with the latter.

The core insight is that the path from legal text to development practice involves translation, and that translation shapes the reality of how companies address the ``Last Mile'' Challenge in AI Governance. Collaborative approaches like the workshop we conducted can support this translation by helping teams discover the improvement-oriented rationale behind requirements themselves. When practitioners understand \textit{why} compliance matters, the quality of their engagement improves.

\section{Evaluating Action}
\label{sec:discussion}

The previous section examined \textit{what} the workshop produced and \textit{how} practitioners' interpretation of requirements shapes governance engagement. This section turns to the knowledge-generation dimension of action research: \textit{how} collaborative engagement functions as a mechanism for governance integration in resource-constrained environments.
Table~\ref{tab:achievements-challenges} summarizes what the workshop achieved alongside challenges that remain unsolved. In the following subsections we reflect on our proposed solution of the ``Last Mile'' Challenge in AI governance.

\begin{table*}[htbp]
    \centering
    \renewcommand{\arraystretch}{1.2}
    \small
    \rowcolors{2}{gray!10}{}
    \begin{tabular}{>{\raggedright}p{1.8cm} >{\raggedright}p{6.6cm} >{\raggedright\arraybackslash}p{6.4cm}}
        \hline
        \textbf{Theme} & \textbf{Workshop Achievement} & \textbf{Open Challenge} \\
        \toprule
        
        Actionable Outputs & 
        Workshop produced actionable outcomes and concrete strategies forming the basis for a 12-month implementation roadmap developed with the CTO & 
        Resource competition with product development remains structural; implementation effort remains substantial \\

        Alignment Discovery & 
        Surfaced latent alignments between compliance and development priorities; identified requirements already satisfied by existing practice & 
        Verification-oriented requirements (e.g., technical documentation) remain disconnected from perceived quality benefits \\
        
        Practitioner Attitudes & 
        Shifted framing from ``necessary evil'' to potential benefit; enabled discovery of alignment rather than top-down persuasion & 
        Durability of attitude shifts under sustained operational pressure remains unknown \\
        
        Ownership \& Expertise & 
        Generated practitioner ownership through collaborative discovery rather than external imposition & 
        Did not transfer compliance expertise; legal interpretation remains specialized work requiring ongoing translation \\
        
        Governance Visibility & 
        Made previously invisible bridging work visible and collective; governance became shared concern rather than isolated burden & 
        Visibility does not create additional capacity; final compliance responsibility remains with AI governance officer \\
        
        \bottomrule
    \end{tabular}
    \caption{Workshop achievements and unsolved challenges, illustrating collaborative workshops as mechanisms for governance integration rather than solutions to underlying structural tensions.}
    \label{tab:achievements-challenges}
\end{table*}

\subsection{The Struggle of Governance Work in Practice}

\textbf{Governance Challenges in a Startup.} The challenge of embedding AI governance in resource-constraint environments is not primarily technical~\cite{nahar2022collaboration, mokander2023operationalising}. It is relational and interpretive: how does one make regulatory requirements meaningful to colleagues who experience them as external impositions competing for scarce development time? Research consistently identifies organizational factors as primary obstacles: harmonizing standards, demarcating scope, and driving communication and change management~\cite{mokander2023operationalising}. These challenges are particularly acute in smaller organizations, where dedicated governance teams, comprehensive toolkits, and extensive consultancy are not feasible~\cite{hopkins2021machine, crockett2021building}. Post-workshop surveys confirmed this structural tension: seven of eight participants agreed that ``AI governance requirements compete with product quality improvements for limited time and resources.'' Even after a collaborative session that surfaced alignment opportunities, practitioners recognized the resource competition.

\textbf{Who Decides how Governance is Operationalized?} The EU AI Act specifies what must be achieved but leaves discretion in how. Currently, much of this determination falls to leadership, governance officers, and legal advisors---or in small businesses without dedicated governance roles, whoever can spare the time. Development teams typically implement what others define. This division of labor may be efficient, but it creates conditions for superficiality. Top-down mandates risk creating the very decoupling between formal policies and actual practices that scholars have documented~\cite{ali2023walking, munn2023uselessness, bietti2020ethics}. Successful implementation requires alignment with existing workflows; innovations ``stick better'' when they sustain rather than disrupt established practices~\cite{sloane2022german}. Our workshop exemplified this principle through what Sloane and Zakrzewski call ``piggybacking''~\cite{sloane2022german}: rather than requesting dedicated time for a standalone governance intervention, the AI Governance Officer proposed the workshop during an already-scheduled two-day strategy session where remote developers came to the company's headquarters for product planning. This alignment with existing rhythms reduced friction and signaled that governance was continuous with product work, not separate from it.

\textbf{Making Governance Work Visible and Collective.} Before the workshop, AI governance existed primarily in the AI Governance Officer's head and in documentation others rarely engaged with. This reflects what Deng et al. describe as ``bridging work''~\cite{deng2023investigating}: the iterative labor required to translate between different perspectives, work that ``teams don't understand'' and that creates ``additional burden, frustration, and burnout''. The AI Governance Officer had been performing this work alone: translating legal requirements into technical language, advocating for governance attention amid product launch pressures. This labor remained invisible precisely because it happened in the spaces between colleagues' primary work. The workshop made this labor visible and collective. For 90 minutes, governance became everyone's work. Governance moved from something done \textit{to} the team to something done \textit{by} the team.

\subsection{The Value of Collaboration}

\textbf{From ``necessary evil'' to Discovered Alignment.} Two post-workshop responses capture a transformation in how participants understood governance:

\begin{quote}
``It is interesting to think about regulatory requirements as a base for possible improvements in our product, instead of being annoyed by them.'' (P4)
\end{quote}

\begin{quote}
``I now see governance less like a necessary evil and do start to see how there can be an overlap with improving product quality.'' (P5)
\end{quote}

The phrase ``instead of being annoyed'' reveals a prior framing of governance as something to endure. The language of ``necessary evil'' positions compliance as fundamentally opposed to productive work: necessary (legally required) but evil (burdensome, value-destroying). The workshop destabilized this binary through discovery rather than persuasion. When the \textit{Performance Evaluation} group recognized that their planned Langfuse implementation for debugging also satisfied logging requirements, this was not information transmitted but connection made. The alignment existed in their work already---the workshop surfaced it. As one participant reflected: ``Before the meeting, I wasn't aware of how much we have in common with the law and that some of the points are also in our own interest'' (P7). This finding resonates with prior research suggesting practitioners are not hostile to regulation and many already implement governance practices without recognizing them as compliance~\cite{koh2024voices}.

\textbf{Discovery Generates Ownership.} Ali et al. document how ethics workers must rely on diplomatic persuasion because they lack formal authority, taking on ``personal risk'' when advocating for governance~\cite{ali2023walking}. The workshop inverted this dynamic. Rather than the governance officer advocating for compliance and hoping teams would comply, participants themselves identified what needed doing and why it mattered. When asked about their preferred governance model, four participants preferred ``Collaborate'' (governance officer and developers decide and implement together) and two preferred ``Involve'' (implement together). None preferred purely informational involvement. The strategies generated were not technically novel. Any compliance consultant might recommend implementing logging, adding AI disclosures, documenting oversight mechanisms. What differed was ownership. When developers generated strategies from their own assessment of requirements and priorities, the resulting commitments carry different weight than externally imposed mandates.

\textbf{Open Challenges.} Despite the workshop's achievements the right column in Table~\ref{tab:achievements-challenges} warrants equal attention. The workshop did not resolve structural resource competition nor did developers become governance experts: translating legal requirements into accessible terms remains specialized work that someone must perform. Practitioners still feel that verification-oriented requirements do not result in product quality improvements. Collaborative workshops may offer a mechanism for governance integration, not a solution to the underlying tensions between regulatory demands and development priorities.

\section{Limitations}
\label{sec:limitations}

\textbf{Workshop Design Limitations.} Participant feedback identified areas for methodological refinement. Several participants noted that legal terminology remained challenging despite pre-processing requirements for accessibility. One observed that requirements ``could mean everything and nothing,'' indicating that even simplified legal language creates interpretive difficulty. Suggestions included clearer task instructions for the strategy ideation phase and more explicit guidance on requirement scope and complexity.
The small group structure was necessary to enable detailed discussion of all 14 requirements within the available time. However, it meant each group had limited understanding of the whole process and all requirements, resulting in some overlap and occasionally mismatched strategies. Future iterations might benefit from structured cross-group synthesis.

\textbf{Limits of a Single Intervention.} We should not overstate what a 90-minute workshop can achieve. The transformation observed in survey responses may prove ephemeral. Organizational pressures like product timelines, competing priorities, or the invisibility of governance labor, do not disappear because of one collaborative session.
Moreover, not all requirements align with development priorities. Technical documentation, rated as high-effort with primarily compliance benefit, represents genuine burden without clear quality payoff. Honest governance integration must acknowledge these tensions rather than dissolving them through rhetorical reframing.
Our study tracked implementation over eight weeks. Longer-term tracking continues, but sustained engagement requires more than initial enthusiasm. Future work should examine whether collaborative practices persist, what organizational factors predict sustained engagement, and how governance approaches adapt as regulatory interpretation evolves.

\textbf{Generalizability Constraints.} Our findings report one case study of a resource-constrained startup with proactive governance orientation and existing ISO certifications; results may differ in less favorable conditions. The first author's insider positioning enabled access to implementation processes that external researchers rarely obtain~\cite{scheuerman2024walled}, but creates inherent tensions: participants may respond differently to a colleague than to an external researcher. We addressed this through external co-facilitation, academic supervision, and transparent positioning, but cannot eliminate insider dynamics.
These constraints represent trade-offs inherent to in-depth qualitative research in real-world AI development. We provide detailed description of context, methods, and findings to enable assessment of transferability. We encourage further research applying collaborative governance approaches in different organizational settings and regulatory environments.

\section{Future Work}
\label{sec:future-work}

Several directions emerge from this research, informed both by our findings and by the open questions they surface.

\textbf{Sustaining Collaborative Governance.} Our study captures the initial effects of a single 90-minute intervention. Longitudinal research tracking governance practices over months or years would clarify whether collaborative engagement produces durable change or whether initial enthusiasm fades under operational pressures. Of particular interest is understanding what organizational factors predict sustained engagement: leadership commitment, integration with existing development practices, or visible results from early implementation may all play a role. Comparative studies across organizations with different sizes, sectors, and governance orientations would help identify boundary conditions for collaborative approaches.

\textbf{Addressing Disconnected Requirements.} Our Pattern~3 findings identify two dynamics within perceived disconnection. For requirements experienced as burdensome paperwork in fast-paced development environments (technical documentation, formal risk management), future work could explore whether alternative framings that emphasize system improvement over external verification shift practitioner engagement. For requirements facing legal ambiguity (data governance for foundation model users, awaited SME templates), sector-specific guidance and worked examples may prove more effective than general regulatory text. Both dynamics warrant investigation across different regulatory requirements and organizational contexts.

\textbf{Aligning Team Strategies with Regulatory Intent.} Collaborative workshops empower teams to generate implementation strategies but do not guarantee regulatory alignment. Future research should examine how to integrate legal validation into collaborative processes without reintroducing the top-down dynamics that collaborative approaches seek to avoid. As EU AI Act implementation guidance matures, including anticipated Harmonized Standards and SME-specific simplifications, research should examine how evolving regulatory clarity affects practitioner engagement with governance requirements.

\section{Conclusion}

This paper asked how collaborative workshops can empower technical teams to integrate AI governance requirements into existing software development workflows. Through insider action research embedded within an AI startup, we developed and tested a legal-text-to-action pipeline that translates EU AI Act requirements into concrete actions. Our approach addresses the ``Last Mile'' Challenge in AI governance by involving software developers in the ideation and prioritization of implementation strategies to find interest alignment and foster meaningful AI governance adoption.

We find three patterns how practitioners engage with regulatory requirements: (1) compliance aligns with development priorities, (2) current work already satisfies requirements, or (3) requirements are perceived as administrative overhead.
Practitioners assess requirements serving end-users or their own development needs as meaningful, but might treat verification-oriented requirements as box-ticking exercises. While external oversight of self-interested organizations is essential, distinguishing requirements that drive system improvement from those that primarily document compliance may help focus implementation effort where it creates most value.

Internal expert collaboration offers a practical mechanism for transforming governance from external imposition to shared ownership. The workshop made previously invisible bridging work visible and collective: governance changed from something done \textit{to} the team to something done \textit{by} the team. When practitioners themselves identify connections between regulatory obligations and existing priorities, the resulting strategies carry different weight than externally imposed mandates. Structural resource competition remains, and not all requirements align with development priorities. Yet collaborative engagement achieves something valuable: surfacing latent alignments, distributing governance awareness, and creating conditions where genuine compliance becomes more likely than performative documentation.

\newpage

\section*{Acknowledgements}

We are grateful to the eight workshop participants whose engagement made this research possible as well as to the CTO and leadership of the organisation that allocated time for the workshop and supported the implementation work that followed. Naira Paola Arnez Jordan co-facilitated the workshop and contributed independent observation notes that shaped our analysis; her earlier input on participatory methodology also informed the research design. We thank Dalia Ali and Chiara Ullstein for guidance and methodological feedback on the research design. We also thank the anonymous FAccT reviewers for their reviews that strengthened the manuscript.

\section*{Ethical Considerations Statement}

Workshop participation was a company activity during paid working hours. Contribution of data to the research was voluntary, with informed consent obtained and documented within the pre-workshop survey. Participants were explicitly informed of the first author's dual role as AI Governance Officer and researcher, and that declining research participation would have no impact on their employment or performance evaluation.

Given the small sample size and the first author's insider position, maintaining complete anonymity for survey responses was not possible. We therefore adopted confidentiality-based protocols: individual survey responses would not be shared with the employer, only aggregated findings would be reported, and publications would use pseudonyms where appropriate. The workshop itself involved face-to-face discussion among colleagues and could therefore not be anonymous.

To address potential bias from insider positioning, an external researcher co-facilitated the workshop and provided independent observation. Research design was reviewed by the second author, a senior researcher independent of the company. The company reviewed this manuscript solely to verify that no proprietary or commercially sensitive information was disclosed; the company did not review, edit, or prevent publication of any research findings or interpretations, including those identifying implementation challenges or limitations.

\section*{Positionality Statement}

The first author conducted this research from within the organization it examines, occupying simultaneously the role of AI Governance Officer responsible for implementing compliance work and the role of researcher studying that same work. This position enabled a depth of access that external researchers rarely obtain, including observation of preparatory conversations with leadership and direct involvement in the implementation work that followed the workshop. It also shaped what could be seen. Coghlan \cite{coghlan2019doing} calls this pre-understanding: the accumulated assumptions that organizational membership produces, which can obscure patterns an external researcher might more readily notice. The first author cannot fully separate the interpretation of the workshop from the investment of having designed and facilitated it, or from the professional stake in governance work succeeding at the company.

Having no affiliation with the studied organization, the second author provided an external academic perspective throughout. Weekly supervisory meetings served as a recurring check on insider framings, surfacing questions the first author had not considered and pushing back on claims that risked overstating the intervention's effects. This arrangement does not dissolve the tensions inherent to insider action research, but it introduces a vantage point outside the organizational culture and commercial context of the research site. Our findings should be read with this positioning in mind.

\section*{Generative AI Usage Statement}

Generative AI tools were used during the preparation of this manuscript. Specifically, Claude (Anthropic, versions Sonnet 4.5 and Opus 4.5) assisted with: (1) identifying and summarizing relevant literature during the literature review process, (2) grammar and style editing to improve clarity and fluency, (3) formatting of figures and tables, (4) reviewing draft sections to identify potential weaknesses and areas for improvement, and (5) drafting the image descriptions for accessibility compliance. All AI-generated suggestions were critically evaluated by the authors, and the authors retain full responsibility for the originality, accuracy, and integrity of all content in this manuscript.

\bibliographystyle{ACM-Reference-Format}
\bibliography{references}

\newpage

\appendix

\section{Detailed Follow-up Tracking of First Actions}
\label{app:tracking}

The three actions reported below were selected to illustrate different relationships between compliance requirements and development priorities: strong alignment (logging), compliance documenting existing practice (AI interaction disclosure), and compliance as additional burden (technical documentation). The reported implementation of the actions happened within 8 weeks after the workshop.

\subsection{Logging System}

\textbf{Action:} Set up Langfuse instance.

Langfuse is an open-source observability platform designed for LLM applications.\footnote{\url{https://langfuse.com/}} It provides tracing, monitoring, and analytics capabilities that allow development teams to inspect inputs, outputs, latency, token usage, and costs across AI workflows. For our context, Langfuse addresses both a development need (debugging AI behavior, understanding customer-reported errors) and a compliance requirement (Article 12 logging for traceability and post-market monitoring).

\textbf{Implementation.} Two full-stack developers implemented Langfuse as a self-hosted instance within the company's cloud infrastructure. The implementation required approximately 32 person-hours across two working days. On the first day, the team encountered a technical roadblock related to the self-hosting configuration. Given competing priorities, the CTO timeboxed the remaining work to one additional day, noting that further delays would risk missing a critical customer delivery deadline. The implementation was completed within this constraint.

This resource competition illustrates that even for strategies with strong compliance-quality alignment, implementation competes with other development priorities. The 32-hour investment represents a substantial commitment for a team of fewer than ten developers.

\textbf{Current scope.} Langfuse is deployed and operational in the development, test, and production environment. AI workflows are instrumented with tracing, enabling the team to inspect LLM calls, monitor performance, and diagnose errors. One AI engineer reported that Langfuse has already proven valuable for customer support: ``Previously when customers reported errors, we relied on their descriptions which were often incomplete. Now we can trace exactly what happened.''

\textbf{Compliance status.} The current implementation partially addresses Article 12 requirements. Logging capabilities now enable recording of events relevant for identifying risks (Article 12(2)(a)) and facilitating post-market monitoring (Article 12(2)(b)). However, a gap remains: Article 12(3)(d) requires identification of natural persons involved in result verification for certain high-risk systems. The current implementation tracks usage at the tenant (organization) level rather than individual user level (human overseer). Addressing this gap would require changes to the authentication and logging architecture and potentially raising conflicting concerns with data privacy.

\subsection{AI Interaction Disclosure}

\textbf{Action:} Audit current AI disclosure touchpoints.

Article 50(1) requires that natural persons interacting with an AI system are informed of this interaction unless it is obvious from context. The audit examined whether current practices meet this requirement.

\textbf{Findings.} The primary AI interaction point is a chatbot feature. The audit identified three disclosure mechanisms already in place: (1) the feature name explicitly references a well-known chatbot (e.g., similar to ``ChatGPT''), (2) a commonly recognized AI icon appears alongside the name, and (3) upon opening the chat interface, an introductory message discloses how the system should be used, notes that AI can make mistakes, advises users to verify results, and states that the user remains responsible for decisions.

Based on this assessment, the requirement was judged to be sufficiently fulfilled by existing practice. No product changes were required.

\textbf{Documentation.} Compliance work consisted of creating a documentation file recording the assessment and verification of Article 50(1) compliance. This documentation provides evidence that the requirement was evaluated and met. The audit and documentation took approximately two hours.

\textbf{Additional considerations.} The audit prompted discussion about extending AI disclosure beyond direct interaction points. Currently, processing steps that involve AI are marked with ``(AI)'' text labels. The team discussed whether adopting consistent visual iconography across all AI-facilitated features would improve transparency, though this was identified as a potential enhancement rather than a compliance gap.

This case illustrates that some requirements may already be fulfilled through existing design practices without explicit legal obligation. The compliance effort was minimal: documenting and verifying what was already in place.

\subsection{Technical Documentation}

\textbf{Action:} Develop C4 architecture diagrams.

Article 11 requires comprehensive technical documentation as specified in Annex IV. Among other elements, Annex IV(2)(c) requires ``the description of the system architecture explaining how software components build on or feed into each other and integrate into the overall processing.''

\textbf{Implementation.} The AI Governance Officer conducted interviews with three software developers and collaboratively created C4 model diagrams at three levels of abstraction:\footnote{\url{https://c4model.com/}} System Context (showing the AI system's relationship to users and external systems), Container (showing major technical building blocks), and Component (showing internal structure of key containers). The collaborative approach ensured accuracy while distributing the knowledge-elicitation burden. Total effort was approximately eight hours.

\textbf{Compliance status.} The diagrams address part of the Annex IV requirements but do not constitute complete technical documentation. Full compliance with Article 11 remains uncertain because the European Commission has not yet released the simplified documentation template for small and microenterprises mandated by Article 11(2). Twenty months after the AI Act's enactment, this template remains unavailable. This regulatory uncertainty creates a practical dilemma: investing in comprehensive documentation now risks economic sunk costs if the simplified template requires a different structure.

\textbf{Perceived value.} When asked about benefits beyond compliance, the developers assessed the value of the documentation as negligible. While technical documentation could theoretically support onboarding, internal communication, or customer-facing trust centers, these potential benefits were not perceived as sufficient to justify the effort independently of compliance requirements. The ongoing maintenance burden---keeping documentation current as the system evolves---was identified as a significant concern.

This case illustrates a requirement where compliance and development priorities show limited alignment. The documentation was created primarily as a compliance exercise rather than to address a felt development need.
\section{Data Collection Protocols}
\label{app:data}

\subsection{Pre-Workshop Survey}

The pre-workshop survey was distributed three days before the workshop via Microsoft Forms. Estimated completion time was 10--15 minutes. Only the consent question was required; all other questions were optional.

\begin{table*}[htbp]
    \centering
    \small
    \renewcommand{\arraystretch}{1.2}
    \begin{tabular}{>{\raggedright}p{0.5cm} >{\raggedright}p{9.5cm} >{\raggedright\arraybackslash}p{4.0cm}}
        \toprule
        \textbf{ID} & \textbf{Question} & \textbf{Response Format} \\
        \midrule
        \multicolumn{3}{l}{\textit{Consent \& Introduction}} \\
        Q1 & Do you consent to the participant information sheet? & Yes/No (required) \\
        Q2 & What's your name? & Free text \\
        Q3 & What comes to your mind when you hear `AI governance'? & Free text \\
        \midrule
        \multicolumn{3}{l}{\textit{Knowledge \& Experience}} \\
        Q4 & Have you previously participated in any AI governance training? & Yes/No \\
        Q5 & If you answered `Yes' in the previous question, please briefly describe the format (e.g., online course, workshop, consultant-led session) and what you remember from it. (1--2 sentences) & Free text (conditional) \\
        Q6 & How would you rate your current knowledge of the EU AI Act? & 5-point ordinal scale \\
        Q7 & How would you rate your current knowledge of ISO/IEC 42001 (AI Management System)? & 5-point ordinal scale \\
        Q8 & Please list any specific AI governance requirements you're aware of that apply to your work. If you can't think of any, write `none'. & Free text \\
        Q9 & How often do you notice or are you influenced by AI governance requirements/processes in your daily work? & 5-point frequency scale \\
        Q10 & Please describe a specific example where AI governance requirements influenced your work or a decision. If you can't think of one, write `none'. & Free text \\
        \midrule
        \multicolumn{3}{l}{\textit{Attitudes}} \\
        Q11a & AI governance requirements compete with product quality improvements for limited time and resources & 5-point Likert + ``I don't know'' \\
        Q11b & AI governance requirements ask us to do things that don't actually improve product quality & 5-point Likert + ``I don't know'' \\
        Q11c & AI governance requirements and product quality improvements are separate concerns that don't affect each other & 5-point Likert + ``I don't know'' \\
        Q11d & AI governance requirements naturally align with our product quality goals & 5-point Likert + ``I don't know'' \\
        Q11e & Meeting AI governance requirements directly improves our product quality & 5-point Likert + ``I don't know'' \\
        Q11f & AI governance requirements help us discover quality problems we wouldn't have found otherwise & 5-point Likert + ``I don't know'' \\
        Q11g & AI governance requirements make our quality practices more systematic and consistent & 5-point Likert + ``I don't know'' \\
        Q11h & Meeting AI governance requirements and showing proof (certification) increase customer trust in our product & 5-point Likert + ``I don't know'' \\
        \midrule
        Q12 & Is there anything else about AI governance the survey did not cover? Do you have any other remarks before the workshop? & Free text \\
        \bottomrule
    \end{tabular}
    \caption{Pre-workshop survey questions.}
    \label{tab:pre-survey}
\end{table*}

\newpage

\subsection{Post-Workshop Survey}

The post-workshop survey was administered immediately after the workshop, before participants left. Estimated completion time was 10 minutes. All questions were optional.

\begin{table*}[htbp]
    \centering
    \small
    \renewcommand{\arraystretch}{1.2}
    \begin{tabular}{>{\raggedright}p{0.5cm} >{\raggedright}p{9.5cm} >{\raggedright\arraybackslash}p{4.0cm}}
        \toprule
        \textbf{ID} & \textbf{Question} & \textbf{Response Format} \\
        \midrule
        \multicolumn{3}{l}{\textit{Basic Information}} \\
        Q1 & What's your name? & Free text \\
        Q2 & What's ONE thing you learned or realized during this workshop? If you can't think of one, write `none'. & Free text \\
        Q3 & Please list any specific AI governance requirements you're aware of that apply to your work. If you can't think of any, write `none'. & Free text \\
        \midrule
        \multicolumn{3}{l}{\textit{Attitudes (identical to pre-workshop Q11a--Q11h)}} \\
        Q4 & Same eight attitude statements as pre-workshop Q11 & 5-point Likert + ``I don't know'' \\
        \midrule
        \multicolumn{3}{l}{\textit{Alignment Evidence}} \\
        Q5 & Please, give ONE specific example from today's workshop where you saw a compliance requirement align with a product quality improvement. Be as concrete as possible. If you can't think of one, write `none'. & Free text \\
        Q6 & Was this workshop a good use of your time? Why or why not? What could be improved? & Free text \\
        \midrule
        \multicolumn{3}{l}{\textit{Responsibility \& Participation}} \\
        Q7 & Based on your experience in today's workshop, who should be primarily responsible for ensuring AI governance is successfully implemented at [company name]? Select the statement that best matches your view. & 5-point spectrum \\
        Q8 & When a new AI governance requirement needs to be implemented at [company name], what level of developer participation should there be? & 5-point IAP2 spectrum \\
        \midrule
        Q9 & Do you have any other thoughts or feedback regarding the workshop or AI governance? & Free text \\
        \bottomrule
    \end{tabular}
    \caption{Post-workshop survey questions.}
    \label{tab:post-survey}
\end{table*}

\newpage

\subsection{Observation Notes Template}

The external co-facilitator documented observations using a structured template organized by workshop phase. The template guided attention to specific dynamics while preserving flexibility for emergent observations.

\begin{table*}[htbp]
    \centering
    \small
    \renewcommand{\arraystretch}{1.2}
    \begin{tabular}{>{\raggedright}p{2.5cm} >{\raggedright}p{4.0cm} >{\raggedright\arraybackslash}p{7.5cm}}
        \toprule
        \textbf{Phase} & \textbf{Observation Focus} & \textbf{Guiding Questions} \\
        \midrule
        Phase 1: Introduction \& Requirements (15 min) & Initial reactions to AI Act presentation & Engagement, skepticism, confusion, or interest? \\
        & Language in participant questions & Burden vs. opportunity framing? \\
        \midrule
        Phase 2: Status Quo Assessment (20 min) & Group dynamics during Miro documentation & Collaborative brainstorming vs. individual contributions? \\
        & Task comprehension & Is the task clear? \\
        \midrule
        Phase 3: Strategy Ideation (20 min) & Compliance-quality connection & Do participants naturally connect compliance to quality, or struggle to see alignment? \\
        & Self-censoring despite instructions & If yes, intervene: `imagine unlimited budget/time' \\
        & Solution character & Creative/ambitious solutions vs. safe/obvious incremental fixes? \\
        \midrule
        Phase 4: Discussion \& Prioritization (20 min) & Participation patterns & Who drives conversation? Who remains quiet? \\
        & Impact definition & How is `impact' defined and reasoned about? Quality improvement mentioned? \\
        & Decision dynamics & Consensus patterns vs. conflicts? Deference to technical experts or leadership? \\
        \midrule
        Cross-cutting observations & Quotes \& language & Statements revealing compliance-quality alignment (or disconnection) \\
        & Dual role dynamics & Instances where first author's operational role might influence research outcomes \\
        & Unexpected patterns & Connections participants make that weren't anticipated \\
        \bottomrule
    \end{tabular}
    \caption{Observation notes template used by external co-facilitator. The template structured systematic observation across workshop phases while allowing documentation of emergent findings.}
    \label{tab:observation}
\end{table*}

\end{document}